\documentclass[letterpaper, 10pt, conference]{ieeeconf}  

\IEEEoverridecommandlockouts                              

\usepackage{url}
\usepackage{cite}
\usepackage{xcolor}
\usepackage{graphicx}
\usepackage{wrapfig}
\usepackage[ruled,vlined]{algorithm2e}
\usepackage{algorithmic}

\usepackage{amsmath}
\usepackage{amssymb}
\usepackage{amsopn}
\usepackage{multirow}
\usepackage{rotating}

\usepackage{color}
\usepackage[normalem]{ulem}
\usepackage{listings}

\usepackage{subfigure}

\usepackage[colorlinks=false,pdfborder={0 1 1},letterpaper=true,pagebackref=true]{hyperref}
\usepackage{indentfirst}
\usepackage{grffile}
\usepackage[utf8]{inputenc}

\usepackage[normalem]{ulem}
\usepackage{breqn}

\title{\LARGE \bf Designing Human-Robot Coexistence Space}

\author{Jixuan Zhi, Lap-Fai Yu and Jyh-Ming Lien$^{1}$%
\thanks{$^{1}$ Zhi, Yu and Lien are with the Department of Computer Science, George Mason University, 4400, University Drive MSN 4A5, Fairfax, VA 22030 USA,
        {\tt\small {\{jzhi,craigyu,jmlien\}@gmu.edu}}}%
        }

\begin{document}

\maketitle
\thispagestyle{empty}
\pagestyle{empty}
\begin{abstract}
When the human-robot interactions become ubiquitous, the environment surrounding these interactions will have significant impact on the safety and comfort of the human and the effectiveness and efficiency of the robot. 
Although most robots are designed to work in the spaces created for humans, many environments, such as living rooms and offices, 
can be and should be redesigned to enhance and improve human-robot collaboration and interactions. 
This work uses autonomous wheelchair as an example and investigates the computational design in the human-robot coexistence spaces. 
Given the room size and the objects $O$ in the room, 
the proposed framework computes the optimal layouts of $O$ that satisfy both human preferences and navigation constraints of the wheelchair.
The key enabling technique is a motion planner that can efficiently evaluate hundreds of similar motion planning problems.
Our implementation shows that the proposed framework can produce a design around three to five minutes on average comparing to 10 to 20 minutes without the proposed motion planner.
Our results also show that the proposed method produces reasonable designs even for tight spaces and for users with different preferences. 
\end{abstract}

\section{Introduction}

More than ever, robots are designed and developed to work with and around humans.
Inevitably, humans and robots in their day-to-day life are going to share a common space.  
While most robots, in particular humanoid robots, are designed to work in the existing environments designed for human activities, 
we envision that the shared space are likely to evolve in the near future to better accommodate and enhance the ever increasing human-robot interaction and collaboration. 
During the decades after personal vehicles were invented, 
we redesigned the roads and streets to adjust to the vehicle size, speed, traffic volume and, more importantly, the behaviors of the drivers in these cars. 
Similarly, as robots are moved from industrial and laboratory settings into our personal spaces, the spaces must also adopt to the robots to increase the humans' safety and comfort and the robot's efficiency.

This paper pioneers the computational design of human-robot coexistence space. 
We propose to formulate the design problem as an optimization problem  subject to constraints from both human's preferences and robot's motion limitations. 
This simple but flexible framework allows us to create multiple design recommendations within a few minutes. 
The key technical challenge is to overcome the bottleneck resulted from the repeated calls to the motion planner during the optimization process. 
From our study, motion planning takes more than 95\% of the computation time.

\begin{figure}[t]
{\includegraphics[width = 0.234\textwidth]{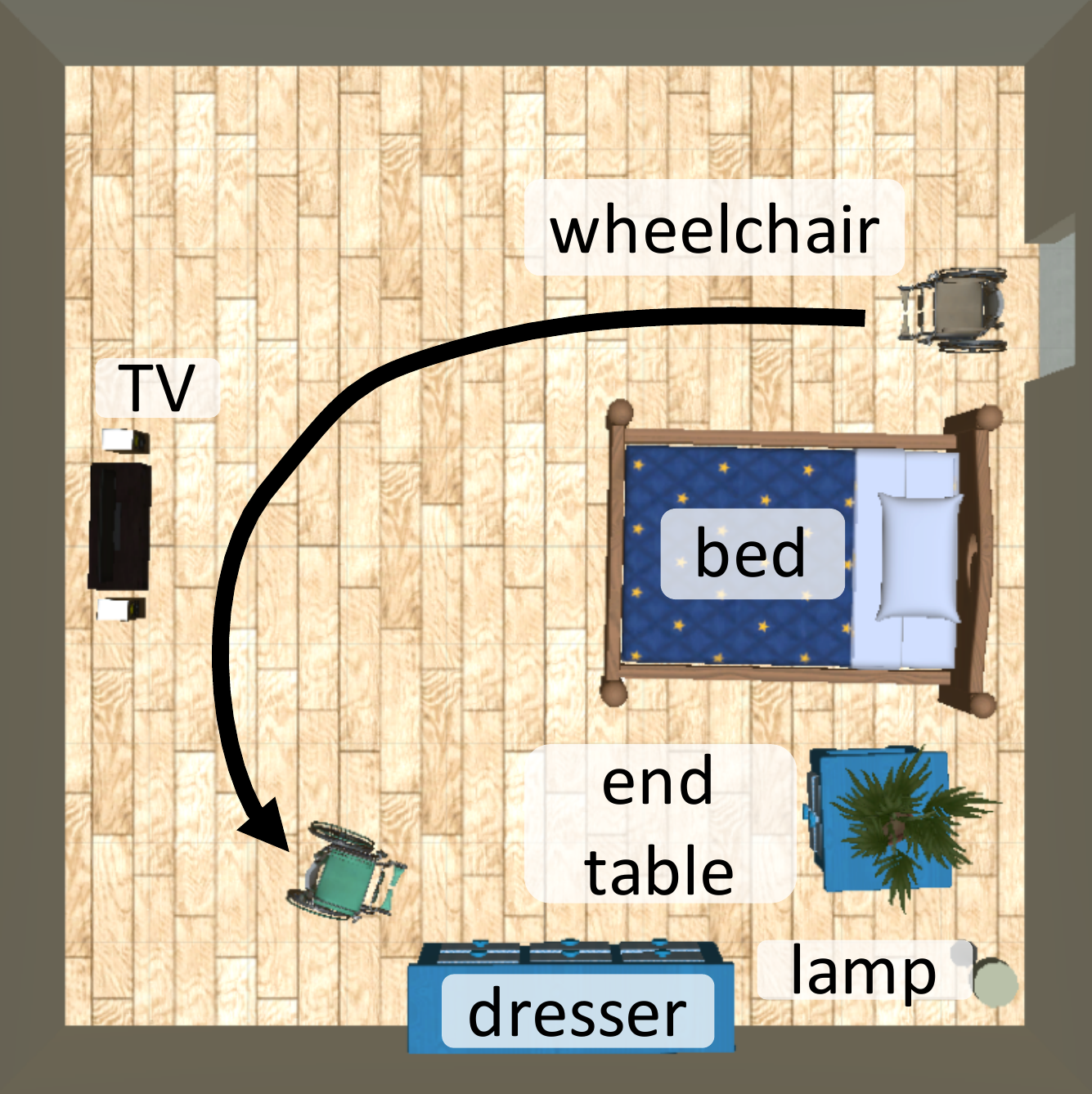}}
{\includegraphics[width = 0.234\textwidth]{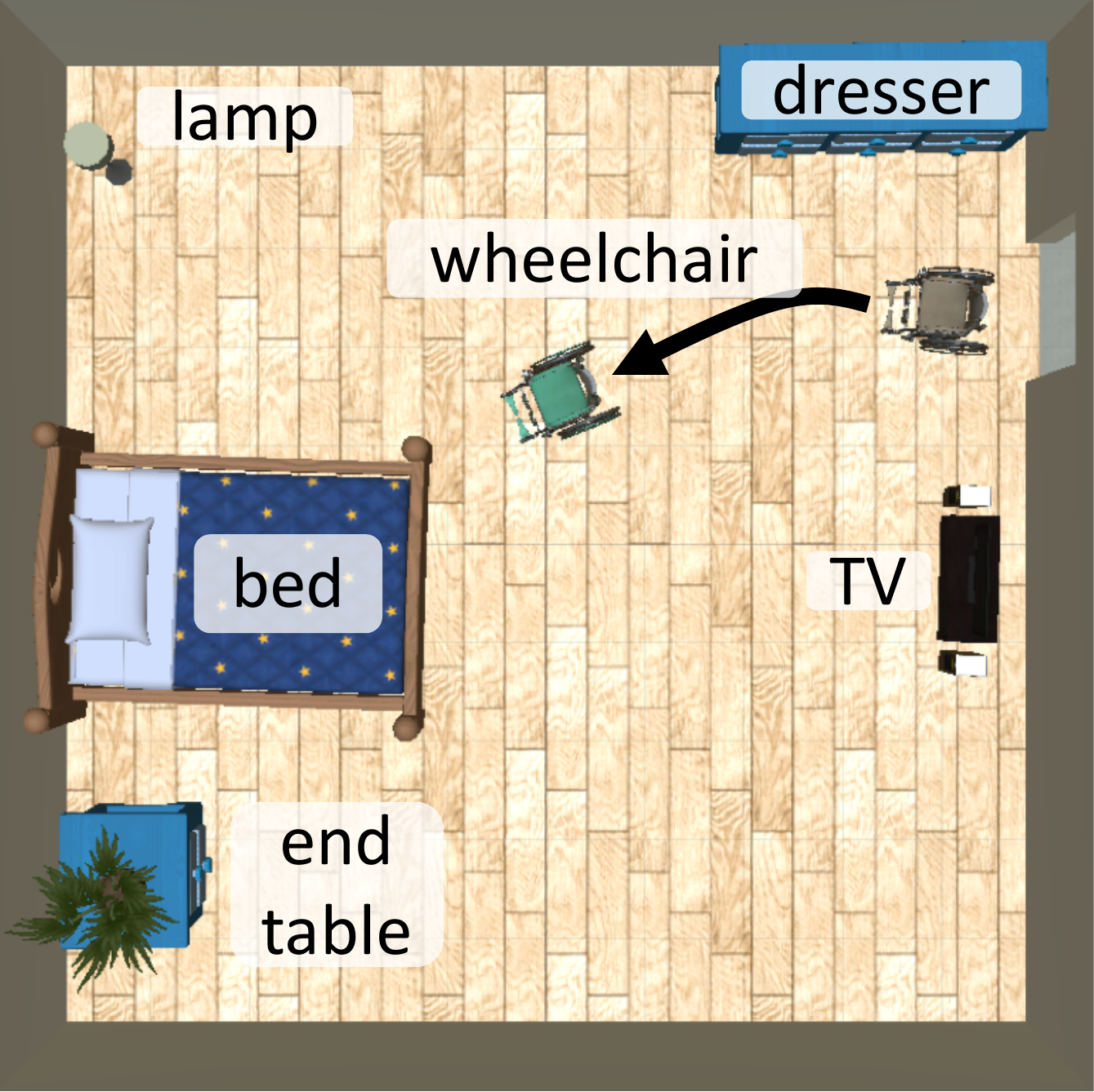}}
\caption{The figure on the left is a bad layout design because the lamp is blocked by the other furniture.
The figure on the right is a better design produced by the proposed framework because all the objects are accessible by the wheelchair robot entering from the door.  }
\label{fig:design-comp}
\end{figure}

In this paper, we focus on designing spaces, such as living rooms and offices, that maximize the accessibility of a wheelchair robot and a person with mobile disability. 
Fig.~\ref{fig:design-comp} shows  the designs with and without the consideration of wheelchair motion.
Although we will use self-driving wheelchair throughout this paper, it should be noted that the proposed framework can be easily adjusted and extended to consider different settings and scenarios. 

\textbf{Main contributions}. This paper presents the first computational method considering the optimization of the space shared by human and robot. 
The paper also contributes the first nonholonomic motion planner that adopts the solutions obtained from earlier motion planning problems to solve more problems in similar but new workspaces.
This new planner can efficiently plan wheelchair motions in hundreds to thousands similar environments in just a few minutes. 

\begin{figure*}[t]
\centering
{\includegraphics[width=0.9\textwidth]{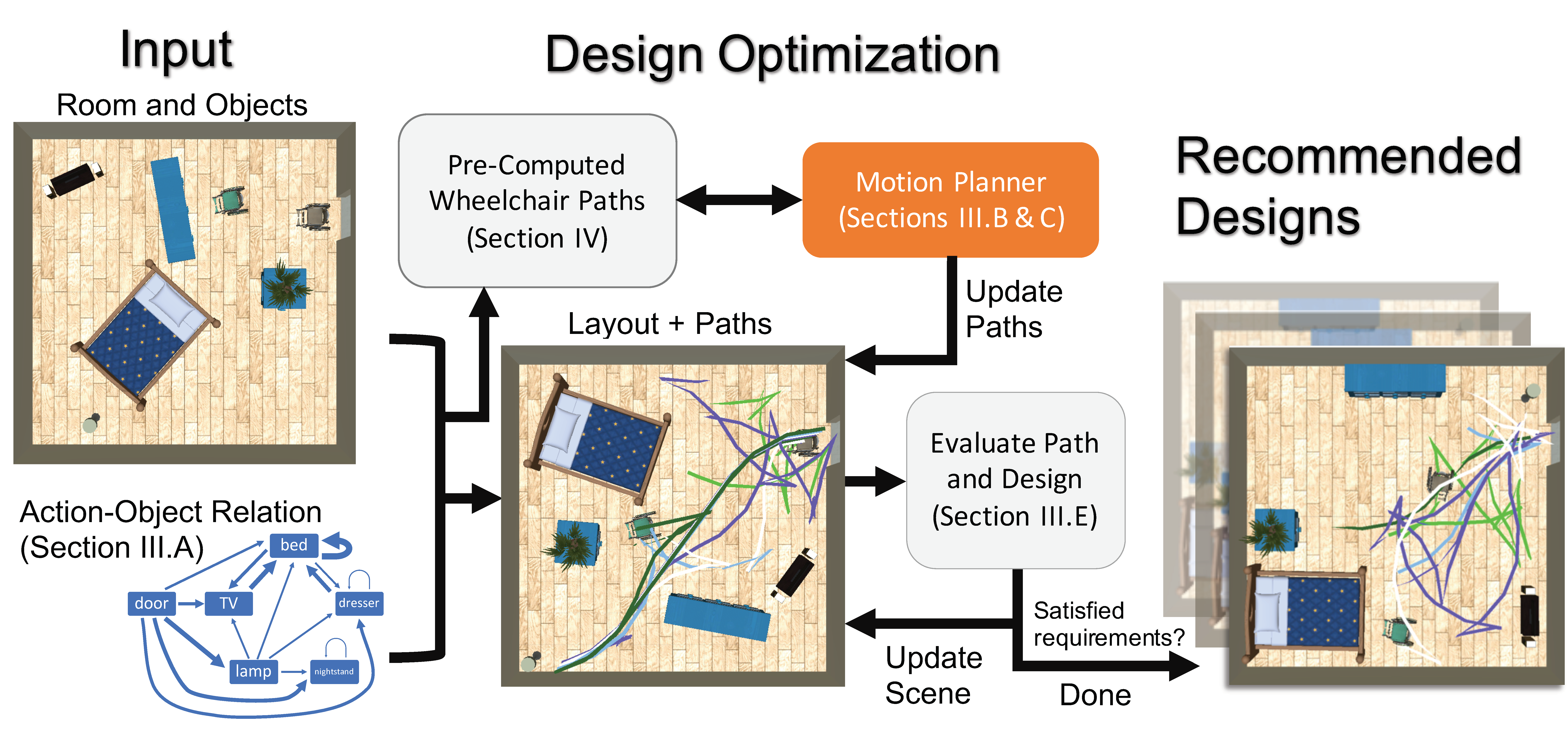}}
\caption{An overview of the proposed design method. Given the room size and objects, with user-defined Action-Object relation graph, our approach generates scenes and paths that satisfy the human preferences and wheelchair motion constraints. }
\label{fig:overview}
\end{figure*}

\section{Related work}

While there has been extensive work on computational design for living spaces, 
there is little effort on designing human-robot coexistence spaces. 
A major technical challenge in designing a human-robot coexistence space is an ultra fast planner that solves motion planning problems in many similar environments. In this section, we will review prior works in both areas.  

\subsection{Computational Design of Space Planning}

Computational design of spaces is usually formulated as an optimization problem. We group these works in rule-based and activity-based methods. 
In this work, we use both rule-based and activity-based representations for human preferences and robot motions, respectively. 

\subsubsection{Rule-based methods}
The first set of methods use user-defined rules to express the preferences of the object placement; for example, the sofa must be against a particular wall and the TV set must face the sofa.
For example, Yu et al.~\cite{yu2011make} proposed to  iteratively change positions and orientations of objects to satisfy these design rules. 
Based on \cite{yu2011make}, Li et al.~\cite{li2020automatic} generated virtual indoor scenarios used for wheelchair training. 
These rules can also be learned from a handful of examples \cite{fisher2012example,xie2013reshuffle}. 
More recently, various neural-network methods \cite{wang2018deep,li2019grains} have been proposed to encode these rules in the latent space from a vast number of examples and have shown to create more complex scenes. 


\subsubsection{Activity-based methods}
These methods synthesize indoor scenes with human-centric inference based on the fact that human activities have a strong effect on arrangement of objects. 
For example, Fisher et al.\cite{fisher2015activity} learned an activity model from a database of 3D models and scenes and, based on the activity model, new scenes are generated.
Several activity models have been proposed, such as activity map \cite{savva2014scenegrok}, action graph \cite{ma2016action},  activity-associated object relation graphs \cite{fu2017adaptive} or 
probabilistic grammar model \cite{qi2018human}. In this paper, an activity model is used to infer the how the wheelchair robot may move around a room.

%
%
%

\subsection{Motion Planning in Similar Environments}

Motion planning methods that utilize previous experiences to solve new problems can be categorized into  space-based methods and trajectory-based methods.

\subsubsection{Space based methods}

Space-based methods learn to plan in similar environments from a representation of the workspace.
For example, Lien and Lu \cite{lien2009planning} 
proposed to construct and store local roadmaps around obstacles, then retrieve 
and merge those local roadmaps into a global roadmap when given a similar new environment. 
Similarly, Experience Graph  \cite{phillips2012graphs, phillips2013anytime,hwang2015lazy} represents the connectivity of a workspace. 
Chamzas et al. \cite{chamzas2019using} used local experiences to global motion planning. They sample and store local maps by local primitives, and then synthesize a global map based on the local maps.

Beyond graph representations, 
Ogay and Kim \cite{ogay2014heuristics} proposed to represent a motion planning algorithm as a random process that is controlled by a heuristic as a distribution of random variables.
A machine learning method then learns to relate the obstacle distribution and the heuristics to better control the motion planner.
Gaussian Mixture Models have been used to represent collision possibility in high dimensional configuration space \cite{huh2017adaptive}.
Ichter et al. \cite{ichter2018learning} proposed to sample configurations from the learned latent space conditioned on the new planning problem. 
Kim et al. \cite{kim2019learning} developed a novel representation of planning problem instance called score-space, based on similarity in score space, they can transfer knowledge from previous problems to solve new task.


\subsubsection{Trajectory based methods}
Trajectory based methods learn to plan in similar environments from recorded trajectories.  For example, Berenson et al. \cite{berenson2012robot} introduced a framework that consists of two parts: one for planning-from-scratch, and the other for retrieving and repairing paths stored in the database.
Saha et al. \cite{saha2016fast,saha2017real,saha2018experience,saha2019real} introduced new machine learning-based algorithms to record a sequence of robot action around obstacles that are reused in more complex environments.
Similar to the space-based methods, Gaussian Mixture Models have also been used to estimate the likelihood of reusing existing trajectories and generate new trajectories in the new environments \cite{barbie2018gaussian}.

When optimality of the trajectory is considered, Hauser \cite{hauser2016learning} proposed the Learning Global Optima (LGO) framework to approximate global optimization that the planner could retrieve and adapt from a library of optimal solutions. 

Although it is developed with a different objective in mind, trajectory prediction
\cite{jetchev2010trajectory,jetchev2013fast} often reuses and generalizes computations obtained from the previous situations.


%
%
%
%
%

\section{Optimization Framework of HR-Space Design}

This section introduces the building blocks used in the proposed design framework. 
Fig.~\ref{fig:overview} illustrates the proposed framework that uses a motion planner and human preferences to evaluate the space layouts.
The framework also uses an action-object relation graph \cite{ma2016action,fu2017adaptive} that encodes the interactions between the human in the wheelchair  and the objects  in the room. 
These interactions will determine the importance of a given trajectory and how much the trajectory should influence the design of the room layout. 
To simplify our discussion, we will use wheelchair to refer to the human in the wheelchair robot for the rest of this paper when the context is clear.

\subsection{Action-Object Relation Graph}
\label{sec:action_model}
\begin{wrapfigure}{r}{0.225\textwidth}
{\includegraphics[width = 0.225\textwidth]{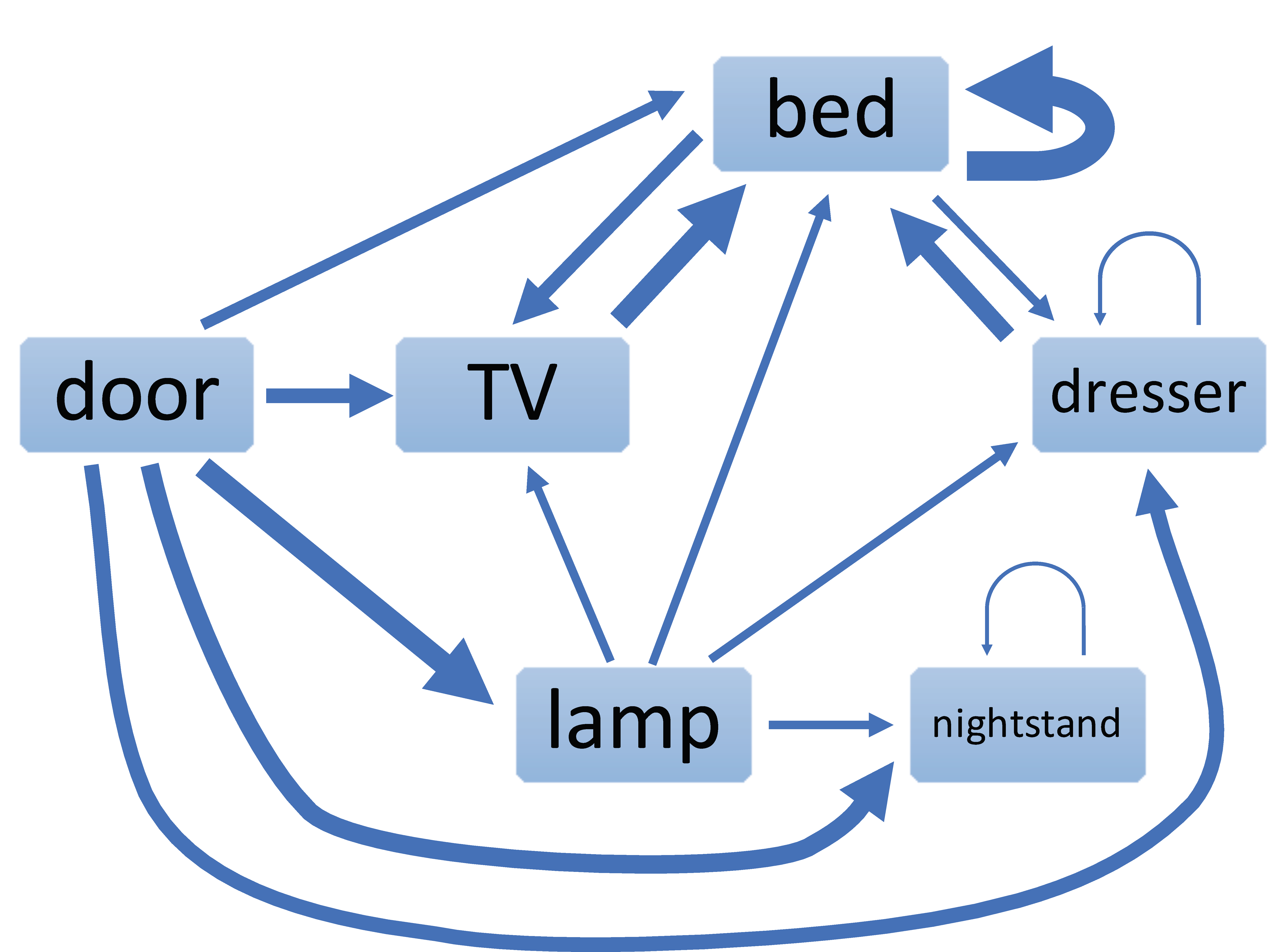}}
\caption{The action object relation graph. Edge thickness represents  the transition probability between two objects.}
\label{fig:actiongraph}
\end{wrapfigure}
In a given room, objects inside the room play an important role in defining the activities of humans and robots. 
These activities in turn define how humans and robots may move around the room. 
Therefore it is natural to produce design recommendations based these object-derived activities. 
To accomplish this, we use {\em action-object relation graph} (AO graph) \cite{fu2017adaptive}. 
As shown Figure~\ref{fig:actiongraph}, an AO graph is similar to the state diagram of Markov chain.
More specifically, an AO graph reveals the activity-associated objects relation. This is a weighted directed graph $G=(V,E)$, the nodes $V$ represent the objects. 
The edges point from one node to another node or a node to itself, and the edge weight is the transition probability from one object to another object or staying still at the current object. 
In this work, the weight measures how likely a wheelchair may move from one object to another. 
Realistic  weights can be estimated from sensors observing human activities \cite{ma2016action,fu2017adaptive}. 


Given an AO graph, we determine the likelihood of the a sequence of objects that the wheelchair may visit. 
That is, from the AO graph, we can determine $M$ most possible sequences that encode different visiting orders of objects in a room. 
Throughout this paper, $M=10$ is used.
Given a sequence $s=\{o_i\}$ of objects  in the room and their corresponding transition probabilities $\{p_i\}$ from the AO graph, 
the likelihood of the path is defined as $L(s) = \Pi_i p_i$.
For each of these $M$ sequences, we use the wheelchair motion planner (detailed in the next sections) to determine if a path exists for visiting all objects in the sequence. 
In Section~\ref{sec:cost}, we will describe how  the information gathered by the motion planner combined with the user-defined preferences
can be used to  evaluate a room layout. 

\subsection{Baseline Wheelchair Motion Planning}
We now sketch the baseline motion planner that
models a wheelchair as  a nonholonomic car \cite{dolgov2008practical}  and uses  a variant of Rapidly-exploring Random Trees (RRT) \cite{lavalle1998rapidly} that 
expands its search using the Reed-Shepp (RS) curves \cite{reeds1990optimal}. 
In Section~\ref{sec:reuse-planner}, this baseline planner will be replaced using a significantly more efficient
planner designed for solving many copies of similar problems. 


The baseline motion planner works as follows: in every step, it first generates a uniformly sampled configuration $(x, y, \theta)$ 
with a biased sampling strategy which has 80\% probability of selecting the goal configuration and 20\% probability of picking the random configuration.
The planner then finds the closest node in the search tree. Instead of Euclidean distance,  the travel distance defined over RS curves is used to determine the closest node. 
For computational efficiency, it is not desirable to calculate the RS curves from the sample to every node in the tree. 
Instead, the planner approximates this by selecting $k$ closest nodes using the {\em weighted} Euclidean distance metric and then finding the nearest 
node with the minimum RS-curve distance among those $k$ nodes. 
The weighted Euclidean distance metric linearly combines the translation in Cartesian space and the change of the orientation between two configurations.  

The planner then  connects the sample to the closest node using the shortest RS curve between them. 
In our implementation, if the target configuration is the goal, the entire curve up to the first invalid configuration is added; 
Otherwise, we add at most $n=5$ steps of the RS curve into the tree. 
Finally, the planner checks whether the newly generated edge is in collision with obstacles. Those obstacles include furniture and walls. 
Only collision-free edges and vertices are added to the tree. The process repeats until the tree reaches the goal. 



\subsection{Touring the Room}

\begin{figure}[ht]
{\includegraphics[width = 0.234\textwidth]{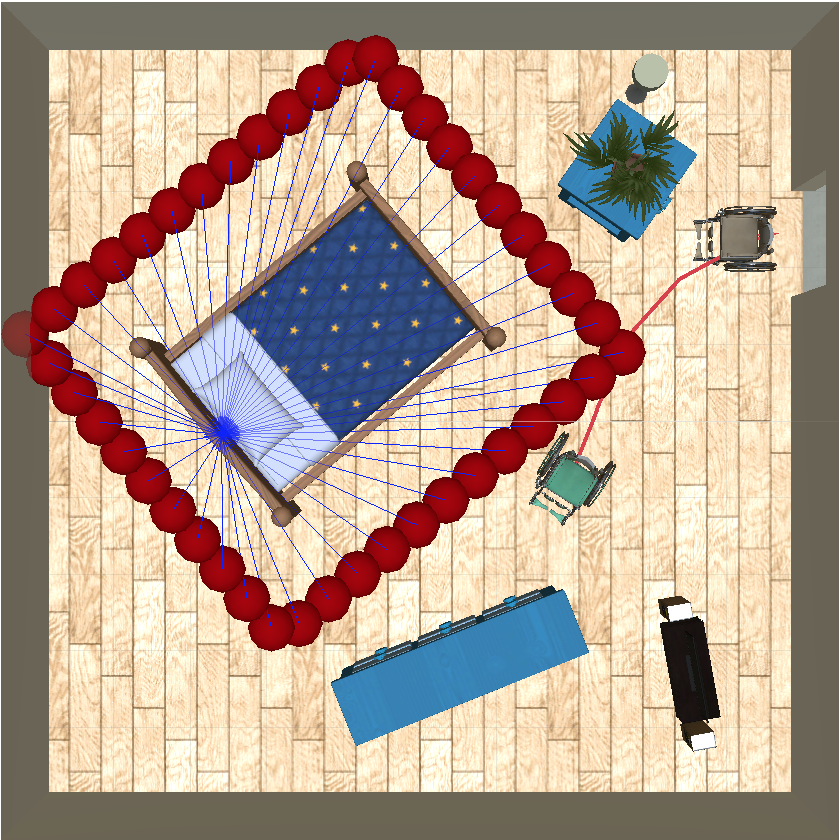}}
{\includegraphics[width = 0.234\textwidth]{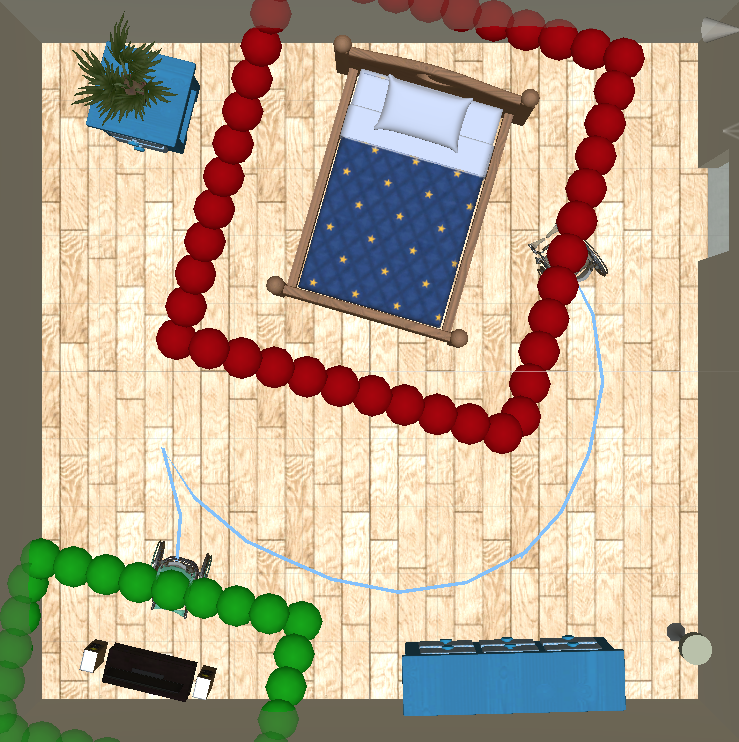}}
\caption{Left: The uniform sampled configurations shown as the red spheres around the bed. 
The blue arrow represents the orientation of the sampled configurations. 
Right: The uniform sampled points around the bed and TV. A valid path in blue between bed and TV can be computed between the sampled points. }
\label{fig:goal}
\label{fig:covered}
\end{figure}

Given a sequence of objects $\{o_i\}$ that the wheelchair  should visit, we now determine a tour that allows the robot to stop by each object. 
To this end, we should determine a sequence of configurations, one for each object, so the path connecting consecutive  configurations in the sequence 
is collision free.
Therefore,  these configurations  should {\em not} be randomly created. In fact,  they should be sampled around the  objects and should be  oriented so that the person sitting in the wheelchair is facing the object. 
To achieve this, 
we uniformly sample $N$ configurations around each object, and each configuration is oriented toward the reference point of   the object. 
The value $N$ is proportional to the perimeter of the object and the size of the wheelchair. 
An example of the sampled configurations around the bed is shown in Fig.~\ref{fig:goal}.

Given the sampled configurations around the objects  $\{o_i\}$ in  a given sequence, such as from the door to TV, dresser and bed in Fig.~\ref{fig:goal}(left), 
we wish to find a collision-free path that visits all objects in that sequence and satisfies the nonholonomic constraint. 
We solve this problem using a branch-and-bound approach. 
To connect objects $o_i$ and $o_j$, all pairs of configurations, one from $o_i$ and one from $o_j$, are sorted based on their RS-curve lengths. 
Starting from the closest pair, the pairs are iteratively evaluated  by the motion planner until a valid path is found. 
Fig.~\ref{fig:covered}(right) shows an example of such a path between the bed and the TV.
Once a path $\pi_{ij}$ between $o_i$ and $o_j$ is determined, we move on to the next pair in the sequence, e.g., $o_j$ and $o_k$.
If we cannot find a path connecting  $o_j$ and $o_k$ by extending $\pi_{ij}$, we  backtrack to $o_i$ and $o_j$ and find another valid (but this time longer) path between $o_i$ and $o_j$. 
This process repeats until a collision-free path visiting $\{o_i\}$ in the given order is found. 

Recall that, in each layout design optimization iteration, the method described in this section must be applied to all $M=10$ sequences obtained from the AO graph. 
As detailed in the next section, these $M$ tours will  be used to evaluate the accessibility of the  design.

\subsection{Evaluating a Layout Design}
\label{sec:cost}
The quality of a design  is encoded as the weighted sum of two cost terms. 
One cost term describes the accessibility of the room, and the other represents how  the placement of the objects
reflects the human preferences.  
More specifically, the quality of a layout design is expressed as $C_{total}$:
\begin{equation}
C_{total}= (\sum_P L(s_P)\mathbf{C_P})\mathbf{w_p^T} + \mathbf{C_Iw_I^T} \ ,
\end{equation}
where $P$ is a path, $\mathbf{C_P}$ and $\mathbf{C_I}$ are vectors of path and object costs, respectively, and 
$\mathbf{w_P}$   and $\mathbf{w_I}$ are vectors of weights. 
While $\mathbf{w_P}$  is partially derived from the provided AO graph, $\mathbf{w_I}$ is defined based human preferences.
Because our optimization frame work considers multiple paths (tours) generated from the object sequences in the AO graph,
the accessibility cost is defined as the sum of all path costs weighted by the likelihood $L(s_P)$, where $s_P$ is  a sequence of objects visited by the path $P$.
Lower $C_{total}$ indicates a better design. Details of the cost terms are described below.


\subsubsection{Path Cost}
\label{sec:pathcost}

The path cost $\mathbf{C_P}=[C_P^l, C_P^r, C_P^c]$ evaluates the length, accumulated rotation, and clearance of the path, respectively. 
A shorter and wide path with fewer turns is considered to be better path than a longer, narrower path with many turns.  
Note that if the path cannot be found, the path cost is a large number based on the room size.
Given a path with $N$ waypoints and the $i$-th waypoint has position $p_i$ and orientation $r_i$, the path cost terms are defined as follows.

\textbf{Path Length Cost $C_P^l$}.
The path length is the sum of the distances between consecutive waypoints along the path. 
Suppose there is a forward path with $M$ waypoints, then the path length is defined as:
\begin{equation}
C_P^l = \sum_{i=1}^{N-1}{||p_{i+1}-p_i||}\ .
\end{equation}

\textbf{Path Rotation Cost $C_P^r$}.
Each waypoint has its orientation information, the path rotation could be computed as the sum of difference of orientation between nearby waypoints. Since we have $N$ waypoints.
\begin{equation}
C_P^r = \sum_{i=1}^{N-1}{||r_{i+1}-r_i||}\ .
\end{equation}

\textbf{Path Clearance Cost $C_P^c$}.
The path clearance cost determines how narrow or wide a path is. 
It is computed as  the mean width of the waypoints.

\begin{equation}
C_P^c = \frac{1}{N} \sum_{i=1}^{N}{(||p_i-q_i|| + ||p_i-w_i||)}\ . 
\end{equation}
where $q_i$ and $w_i$ is the position of object which is closest to the waypoint position $p_i$ on the left and right side. 


\textbf{Path Cost Weight $\mathbf{w_P}$.} 
Generally speaking, the weights $\mathbf{w_P}$ determine how  the path costs can be combined. 
We define the weights $\mathbf{w_P}=\alpha[w_P^l, w_P^r, w_P^c]$, where $\alpha$ is a scaling factor of the user defined parameters 
$w_P^l$, $w_P^r$, $w_P^c$ that express the importance of the path length, rotation and clearance in the optimization process. 
The value of $\alpha$ is determined based on the Action-Object Relation Graph (AO graph).
Recall that, from Section ~\ref{sec:action_model}, we obtained multiple object sequences from the AO graph. 
To define $\alpha$, let us first define the scale factor $\alpha_i$ of $o_i$, which is determined by the likelihood of each sequence and the frequency of $o_i$ appeared in the sequences containing $o_i$. 
More specifically, for each object $o_i$ in a sequence $s$, we count the object occurrences as follows: if $o_i$ is the start or end of $s$, $o_i$ is counted once, otherwise
$o_i$ is counted twice as $o_i$ will appear in two path segments in $s$.

Then the frequency $freq_s(o_i)$ of $o_i$ can be determined by the counts divided by total counts in the sequence $s$. 
In addition, each sequence $s$ also has a likelihood $L(s)$ defined from the AO Graph. 
Combined with the likelihood $L(s)$ of all sequences containing $o_i$, 
the scaling factor $\alpha_i$ of $o_i$ is $\sum_{s \in S} L(s) freq_s(o_i)$, where $S$ is all sequences containing $o_i$. 
Finally, we have $\alpha = \sum_i \alpha_i$. 

\subsubsection{Interior design cost} This cost estimates how the object placement reflects human preferences. 
Therefore, we consider the spatial relationship between objects, including the  spatial relationship between each object  and  its nearest wall. 
Certain objects, such as a TV and a couch, have a pairwise relationship including their mutual distance and orientation. 
Suppose that  there is a pairwise relation set $S$ including all pairs of objects.

\textbf{Pairwise Distance Cost $C_I^d$.}
Every object in the room has a determined distance to its nearest wall and may also have a distance preference from other objects. 
This distance cost determines the spatial relationships between objects.  The target distance can be set by user.  
The cost is defined as:
\begin{equation}
C_I^d = \sum_{\forall {i,j} \in S}({||o_i-o_j||-d_{i,j}})^2\ ,
\end{equation}
where $o_i$ and $o_j$ are the positions of objects $i$ and $j$,  and $d_{i,j}$ is the user-defined target distance. 

\textbf{Pairwise Rotation Cost $C_I^r$.}
Another spatial relationship between objects is pairwise orientation. Each object has a relative orientation to its nearest wall and may have a desired relative  orientation to other objects. 
For example, one might require the TV and the couch to face each other. 
The rotation cost is then defined as: 

\begin{equation}
C_I^r = \sum_{\forall {i,j} \in S}({||\theta_i-\theta_j||-\theta_{i,j}})^2
\end{equation}
where $\theta_i$ and $\theta_j$ are the orientations of objects $i$ and $j$, and $\theta_{i,j}$ is the user-defined target relative angle. 

\subsection{Layout Optimization}

We now are ready to  describe how the proposed method creates a realistic layout by minimizing the costs. 
To this end, we apply a simulated annealing method~\cite{kirkpatrick1983optimization} with a Metropolis-Hasting algorithm~\cite{metropolis1953equation,hastings1970monte}. 
Let us first define a Boltzmann-like objective function for state $(P,I)$:

\begin{equation}
f(P,I) = exp(-\frac{1}{t}C_{total}(P,I))\ ,
\end{equation}
where $P$ is the set of paths touring the room, $I$ is the object layout, $t$ is the temperature parameter, which decreases 
over the iterations. 

In each iteration, our method chooses a move to change the present configuration $(P,I)$  to a new configuration $(P',I')$. There are two types of moves:
(1) change the position of the object, and (2) change the orientation of the object. 
Once the layout $I'$ is updated, we proceed to find new paths $P'$. 
With annealing schedule, the optimizer explores the solution space with large moves in the beginning, and tunes the configuration with small moves in the end. 

The new proposed configuration $(P',I')$ is accepted with a probability based on Metropolis criterion:
\begin{equation}
Prob(P',I'|P,I) = min(1 , \frac{f(P',I')}{f(P,I)})\ .
\end{equation}

In the beginning of each iteration, the optimizer may aggressively accepts any  move, even a bad one,  when the temperature is high. As the temperature decreases, the optimizer is less likely to accept a bad move.

%
%
%
%

\section{Planning Motion in Similar Environments}
\label{sec:reuse-planner}

A major challenge in developing the computational framework proposed in the previous section is the motion planning, 
which, as we will see in the experimental section, takes more than 95\% of the total computation. 
Although there exist planners that exploit similarities among problems, they usually focus on a handful, instead of hundreds, of similar problems, 
and do not consider nonholonomic constraints.  
This section presents several new techniques that maximize computation reuse, thus  the overall 
motion planning cost is significantly reduced.

\subsection{Decompose C-Space using Reed-Shepp Words}
\begin{wrapfigure}{r}{0.2\textwidth}
{\includegraphics[width = 0.2\textwidth]{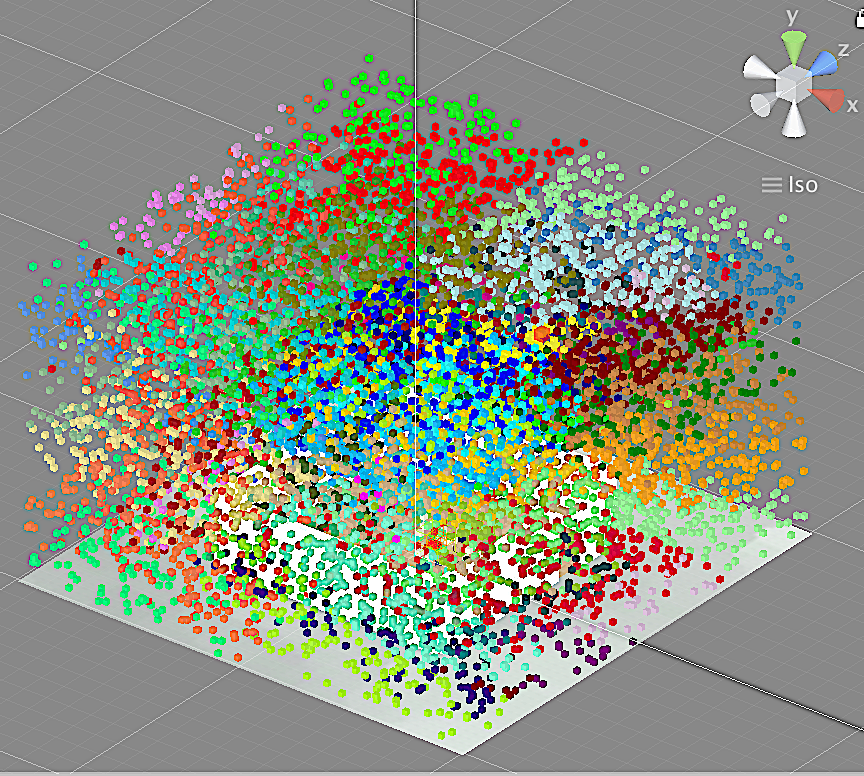}}
\caption{Approximate partition the configuration space into different regions by RS words. Samples with the same color are a region with the shared word. There are in total of 48 regions. }
\label{fig:partition}
\end{wrapfigure}
We propose to decompose the configuration space based on the shortest 
Reed-Shepp (RS) curves with the wheelchair starting at the center of the room. 
Each region in the decomposition is a set of points that share the same RS words. 
Similar decomposition has been studied using analytic methods~\cite{soueres1996shortest,desaulniers1995efficient}. 
However, to the best of our knowledge, these methods remained in theory and 
the only practical implementation is only available for Dubin's car \cite{shkel2001classification}.

We propose to approximate the decomposition via adaptive sampling. We first create $K=1000$ configurations sampled uniformly from the 
configuration space of an empty room that has the size double that of the original room. For each sample $s$, we compute the shortest 
RS-curve and its associated RS-word from the center of the room. We then incrementally add more samples. 
For each additional sample $s$ (also sampled uniformly from the C-space), 
we first find the closest sample $s'$. If $s$ and $s'$ share identical  RS word, then $s$ is discarded. Otherwise, $s$ is added to the samples. 
This process is repeated until 100 consecutive failures are encountered. 
Fig.~\ref{fig:partition} shows the resulting partition of the C-space into 48 regions with the rotational radius of the wheelchair equals to three meters.
We call this partition {\em RS-decomposition}.

When a query configuration $q$ is given, the RS-curve of $q$ can then  be estimated using the RS-word of  $q$'s nearest point in the RS-decomposition.  
With a proper data structure, such as a KD tree, this can be done efficiently by avoiding the evaluation of all 48 cases in determining the shortest RS curve. 
However, because RS-decomposition is only an approximation, the RS-word obtained via the closest sample might be incorrect or even invalid. 
To avoid this, the baseline RRT planner is modified so that the random configurations that pull the tree into unexplored regions are only sampled from the existing 
configurations in the RS-decomposition. We see about 40\% reduction in computation time without noticeable differences in the found paths.

\subsection{Path Reuse}

While RS-decomposition allows us to precompute and reuse RS-curves, it is agnostic to the changes of the designed layout. 
It is also important to know that, the difficulty of  the motion planning problems changes over time as the optimization progresses. 
That is,  when the optimization is just started, the layout tends to be chaotic and changes dramatically, and the problem of finding a tour visiting objects in the scene is difficult.
As the  optimization converges, the layout design stabilizes,  and finding a tour becomes easier. 
In this section, we present ways to utilize this observation and further speed up the motion planner both within 
an optimization iteration and between iterations.

\textbf{Reusing paths within the optimization iteration}.
In each iteration, multiple sequences of objects are evaluated. These sequences share many common trajectories.
Therefore, we reuse these trajectories among the sequences whenever it is possible. 
To achieve this, we maintain a database of trajectories, each of which is indexed by the pair of objects and their corresponding samples that the trajectory connects. 
Note that some trajectories may become invalid due to the displacement of objects in the scene which may result in collision or invalidate the associated RS curves. 
Once detected, an invalid trajectory is removed from the database. 

%


\textbf{Reusing paths from the earlier optimization iterations}. 
As the optimization process converges, the similarity between consecutive optimization iterations increases. This allows the opportunity of reusing paths obtained from the earlier iterations. 
When the motion planner needs to find a path between these objects, it looks up the database built from the earlier iterations and extracts a potential path for the new layout. 
The path is then transformed to the new location and orientation. 
If the transformed path is not feasible, the planner  finds a new path and updates the database.

\subsection{RRT Reuse}

\begin{wrapfigure}{r}{0.24\textwidth}
{\includegraphics[width = 0.25\textwidth]{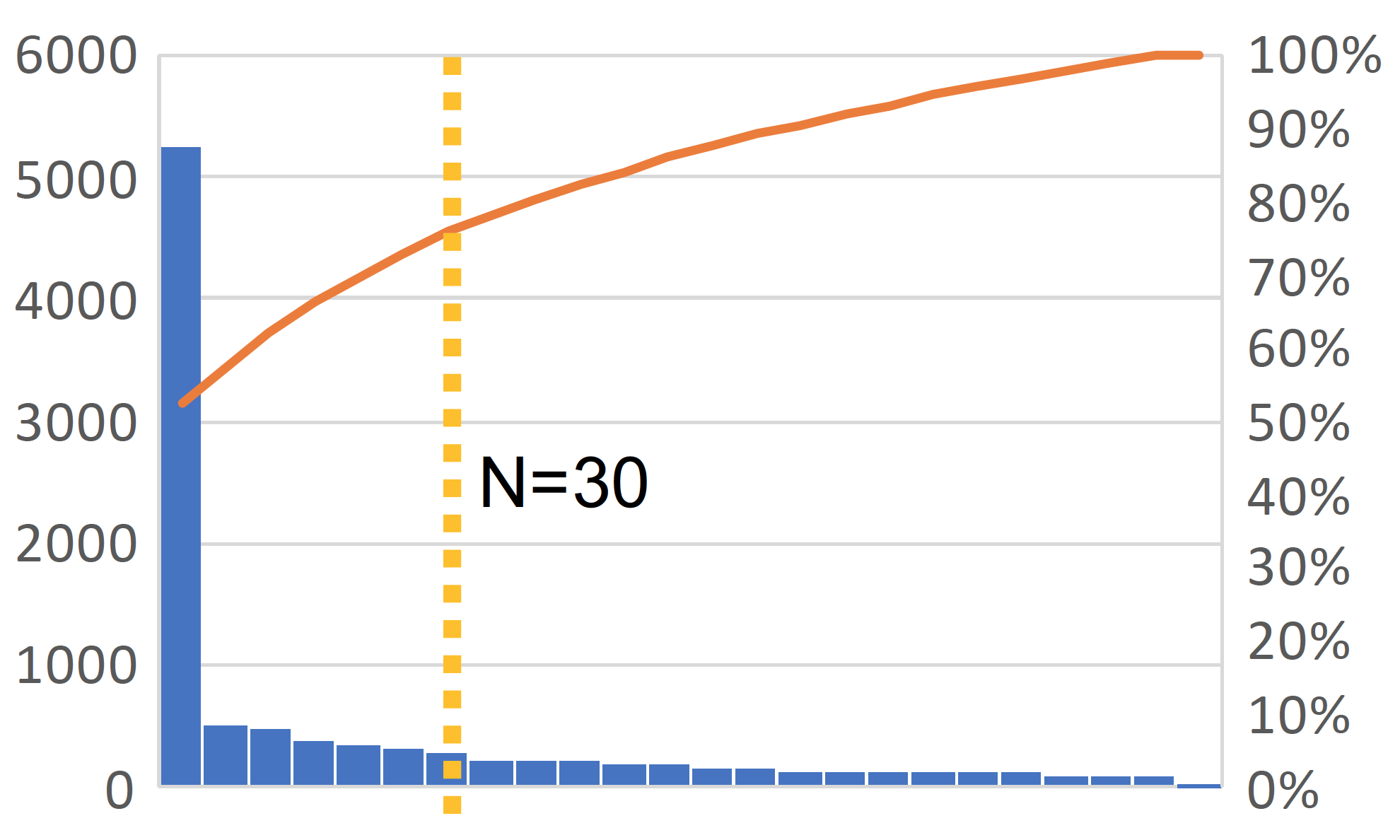}}
\caption{Number of RRT iterations needed for finding a valid path.}
\label{fig:rrt-it}
\end{wrapfigure}
Similar to the case of reusing the paths, as the optimization converges, the changes to the layout decreases. 
Therefore, it is in fact possible that we can reuse a large portion of the tree.
In particular, if the position of RRT root does not change much, we can reuse the tree from the previous optimization iteration. 
From our study shown in  Fig.~\ref{fig:rrt-it}, the number of RRT iterations needed to find a path successfully
 is generally small. Consequently, we do not need many nodes in the reused tree. 
Therefore, if the reused tree grows too large, it may be more time consuming to expand the tree. 
In addition, objects can be displaced in each iteration, and, if the tree is too large, nodes in the tree can easily become invalid. 
Therefore, we rebuild a new tree when the total number of tree node is larger than a user-defined number. In all of our experiments, 30, as indicated in  Fig.~\ref{fig:rrt-it}, is used. 


%
%
%
%
\section{Experiments and Result}
In this section, we report and compare the computation times needed to create a design using the baseline and proposed motion planners. We further study the effects of room size and the AO graph to the designs created by the proposed method. 
We implemented our method on a MacBook Pro laptop with 2.2GHz Inter Core i7 and 16 GB memory. The proposed optimization framework and the  motion planners are implemented in C\# in Unity. 
We demonstrate how our method can be applied in three different room types: bedroom, office, and living room. 
For each environment, we extract 10 most possible sequences from a given AO graph.
The maximum number of optimization iterations in our simulated annealing  is 700 for all experiments.

\subsection{Time Comparison}
\begin{figure}[th]
\centering
{\includegraphics[width = 0.4\textwidth]{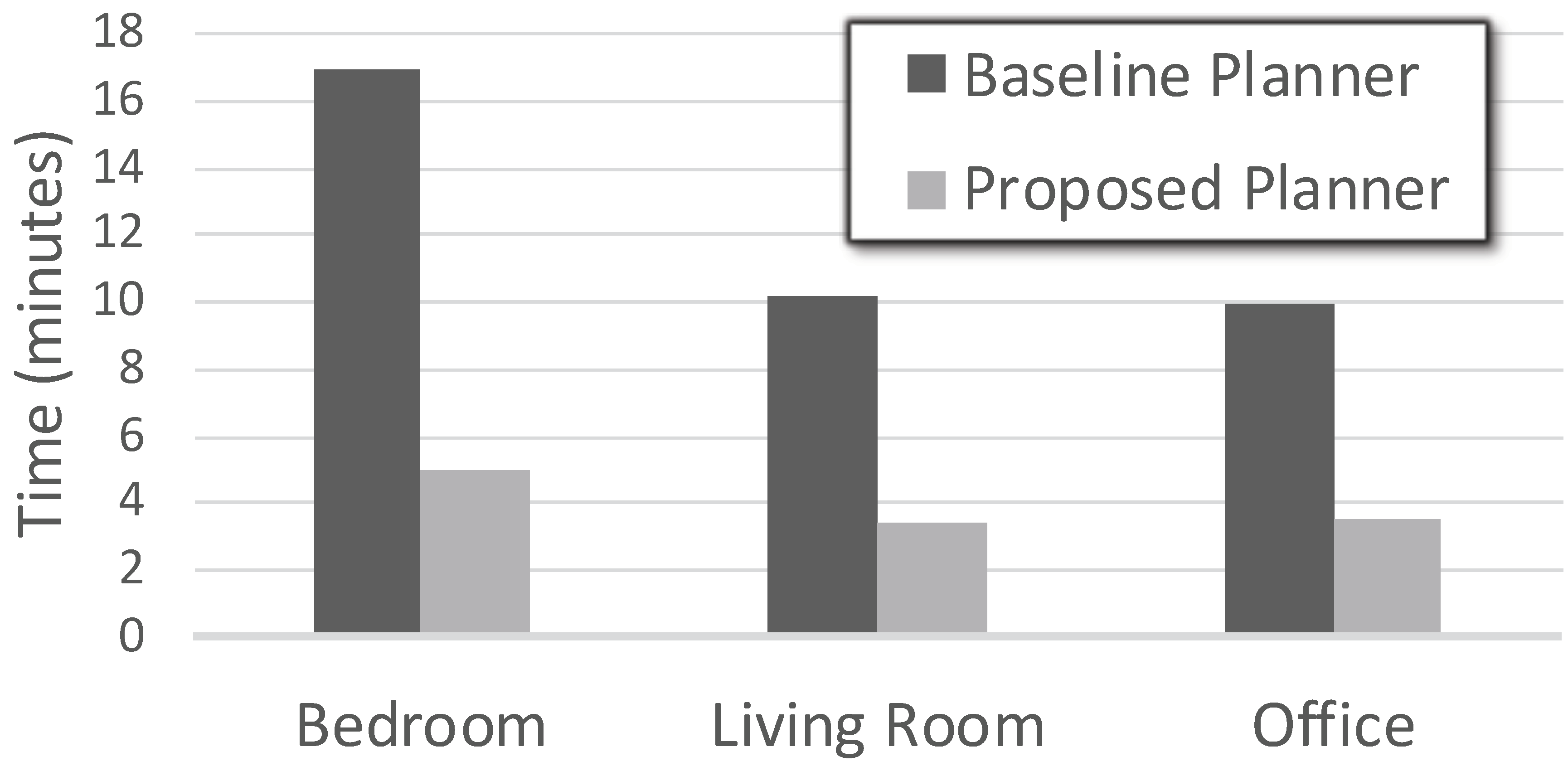}}
\caption{Path planning time using the baseline planner and the proposed planner. The planning time is 96.06\% of the total running times. }
\label{fig:time}
\end{figure}

\begin{figure*}[t]

{\includegraphics[width = 0.24\textwidth]{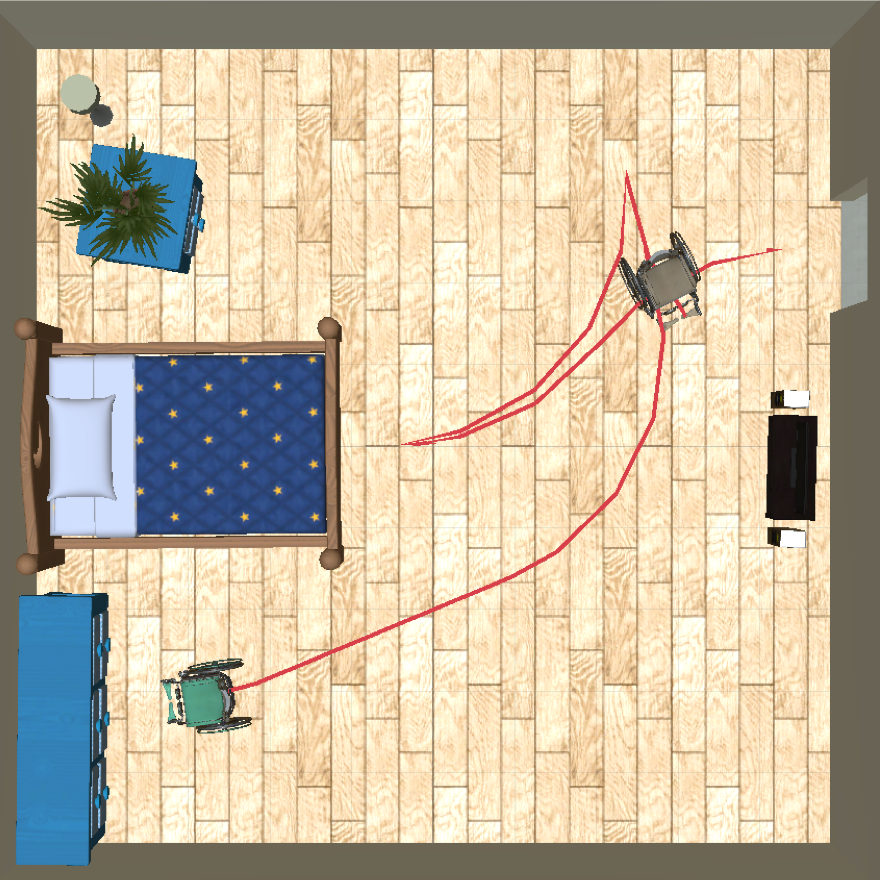}}
{\includegraphics[width = 0.24\textwidth]{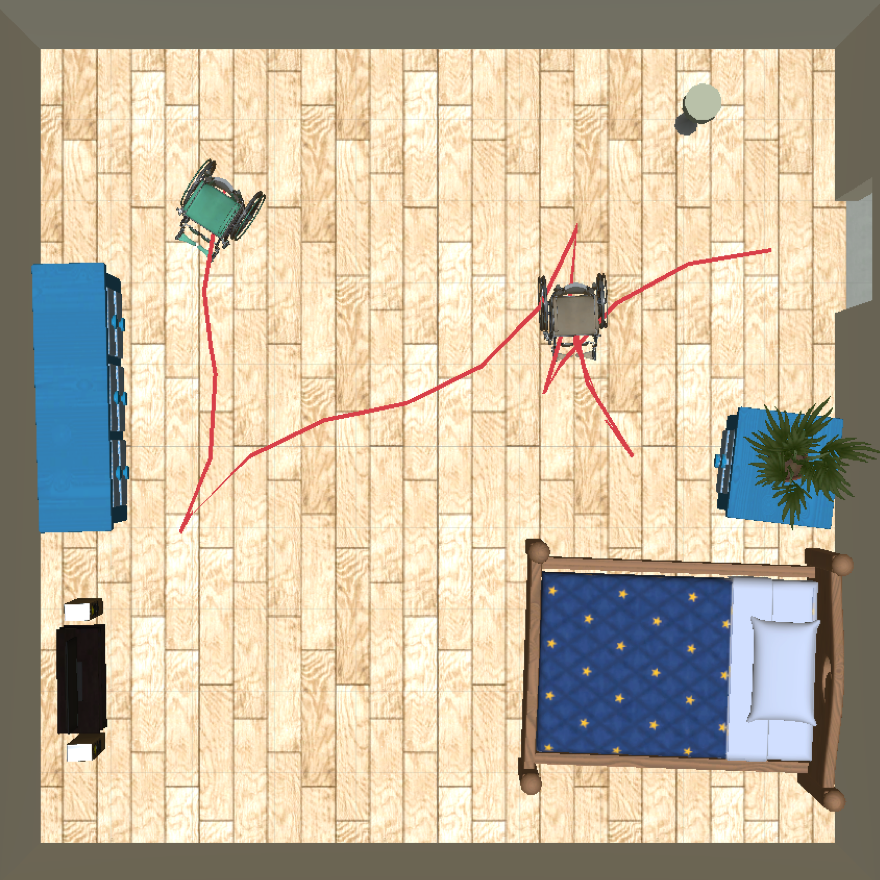}}
{\includegraphics[width = 0.24\textwidth]{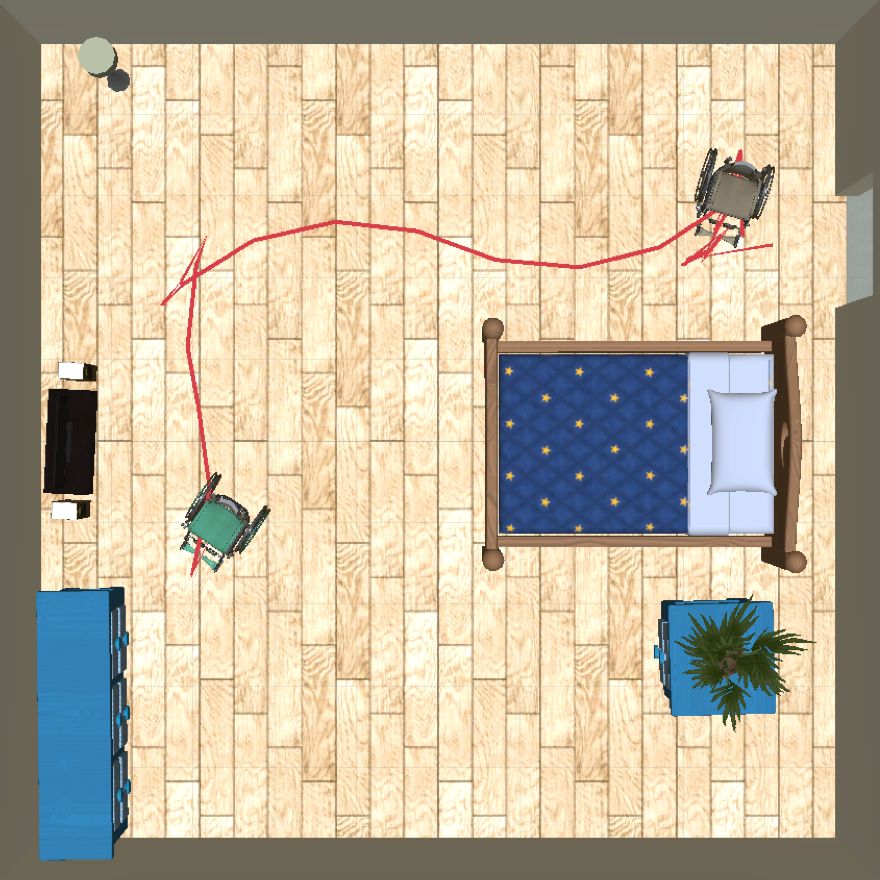}}
{\includegraphics[width = 0.24\textwidth]{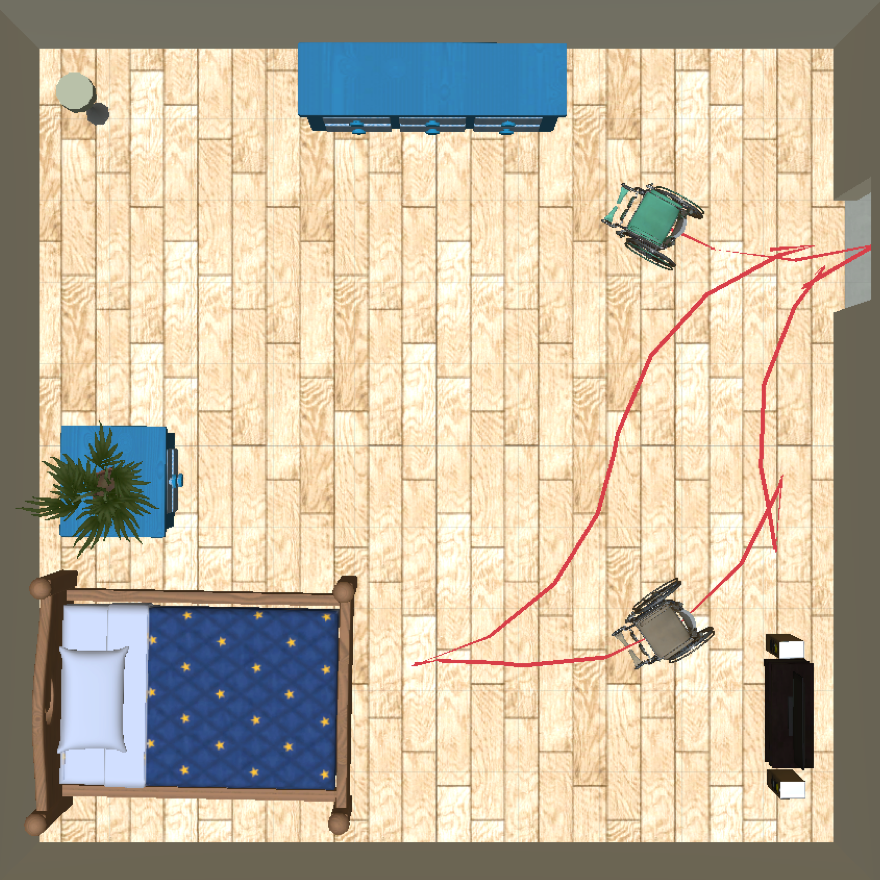}}

{\includegraphics[width = 0.24\textwidth]{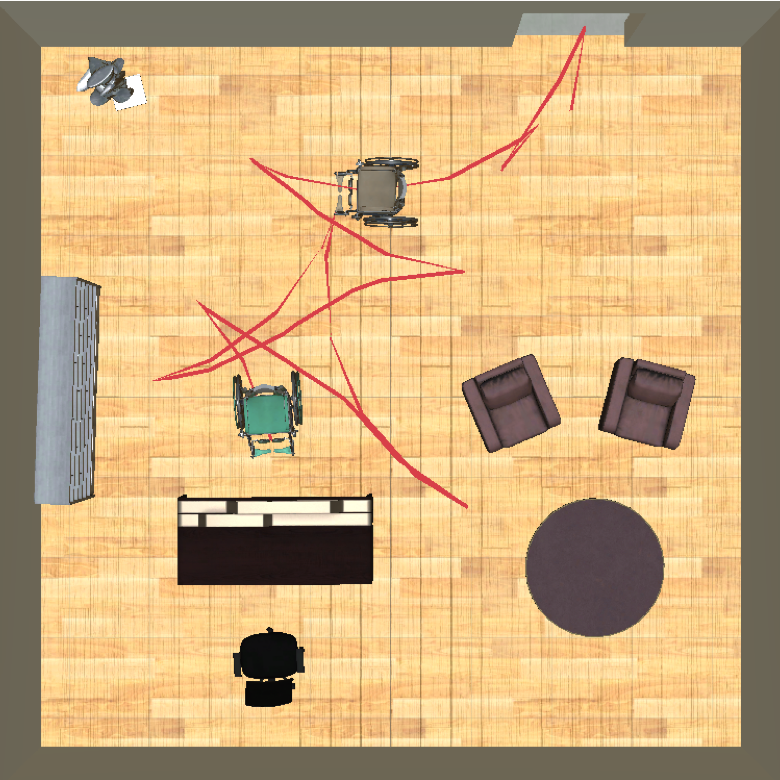}}
{\includegraphics[width = 0.24\textwidth]{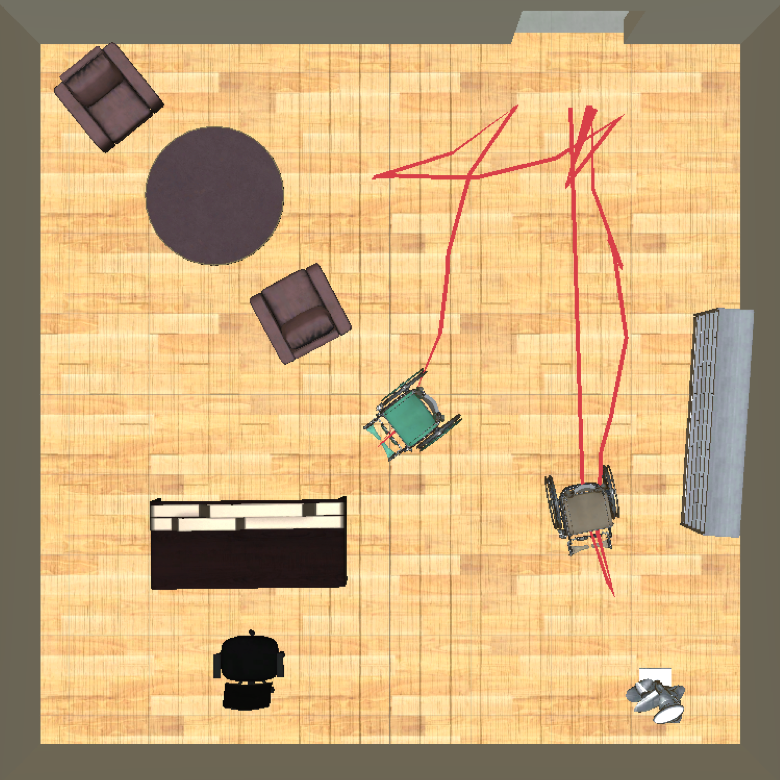}}
{\includegraphics[width = 0.24\textwidth]{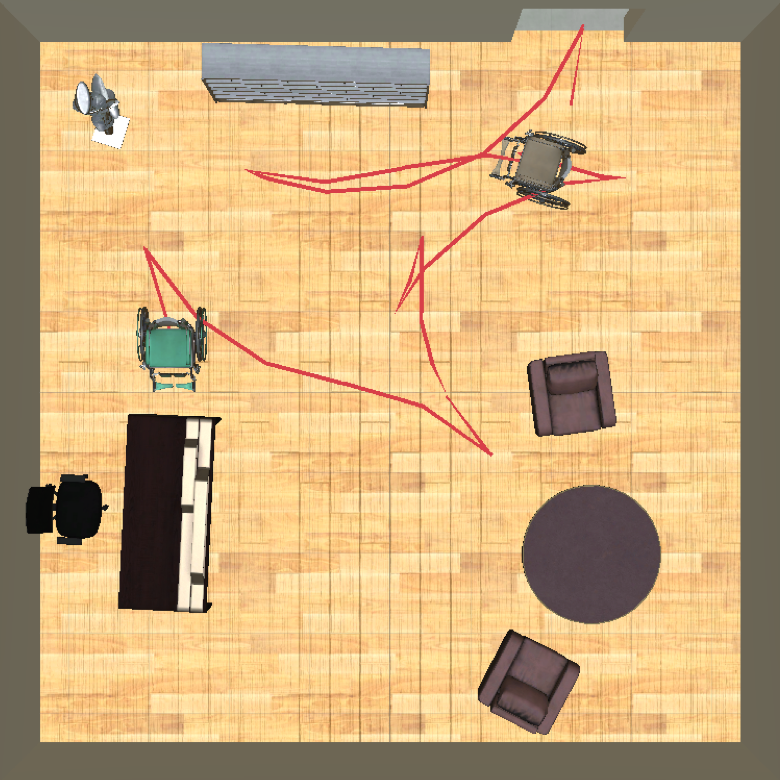}}
{\includegraphics[width = 0.24\textwidth]{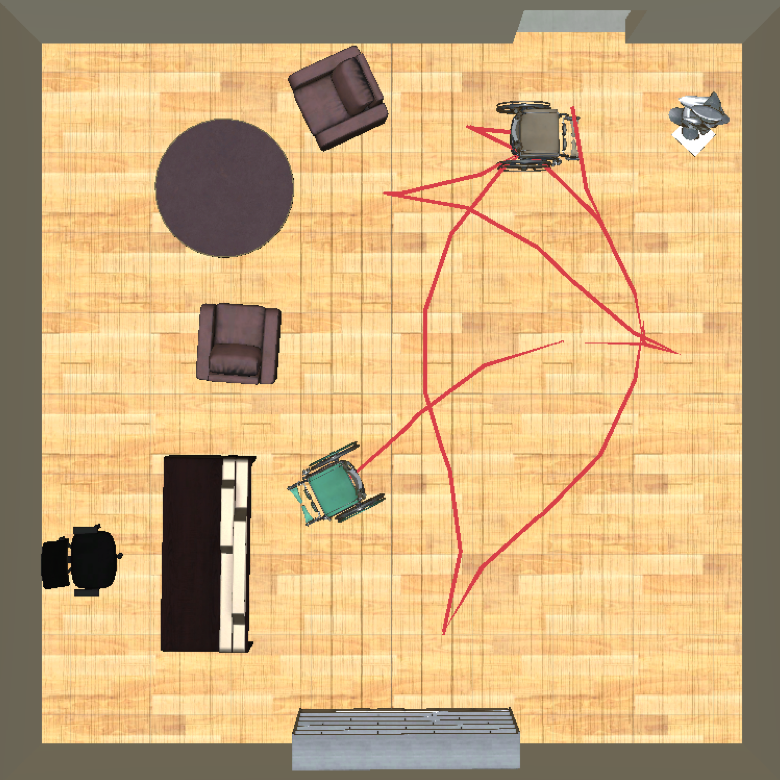}}

{\includegraphics[width = 0.24\textwidth]{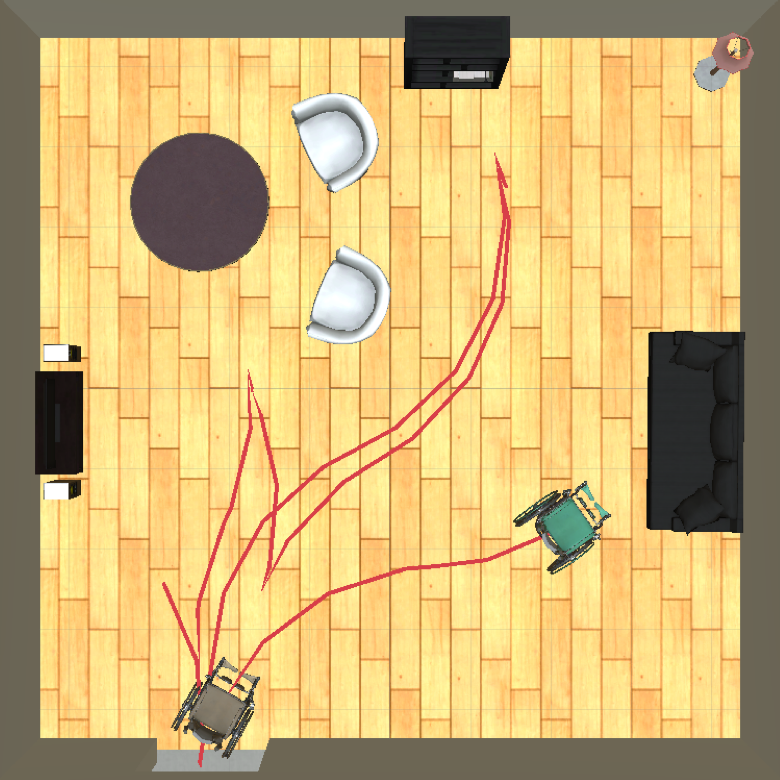}}
{\includegraphics[width = 0.24\textwidth]{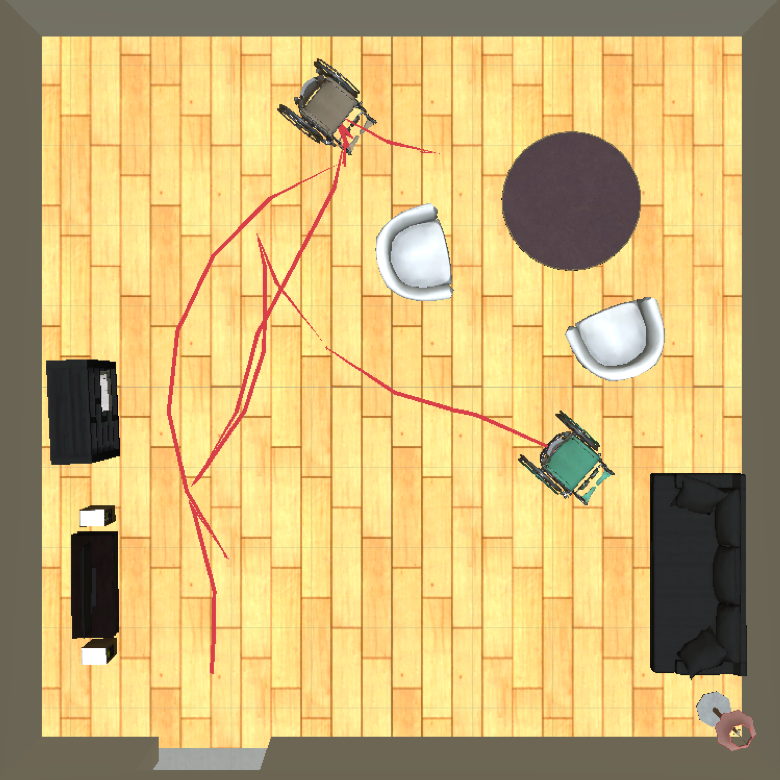}}
{\includegraphics[width = 0.24\textwidth]{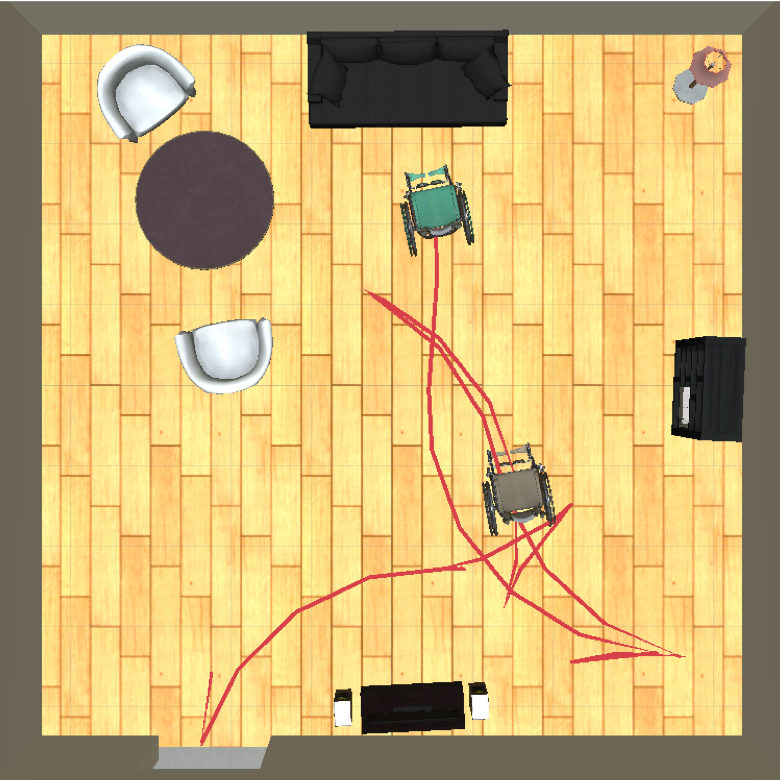}}
{\includegraphics[width = 0.24\textwidth]{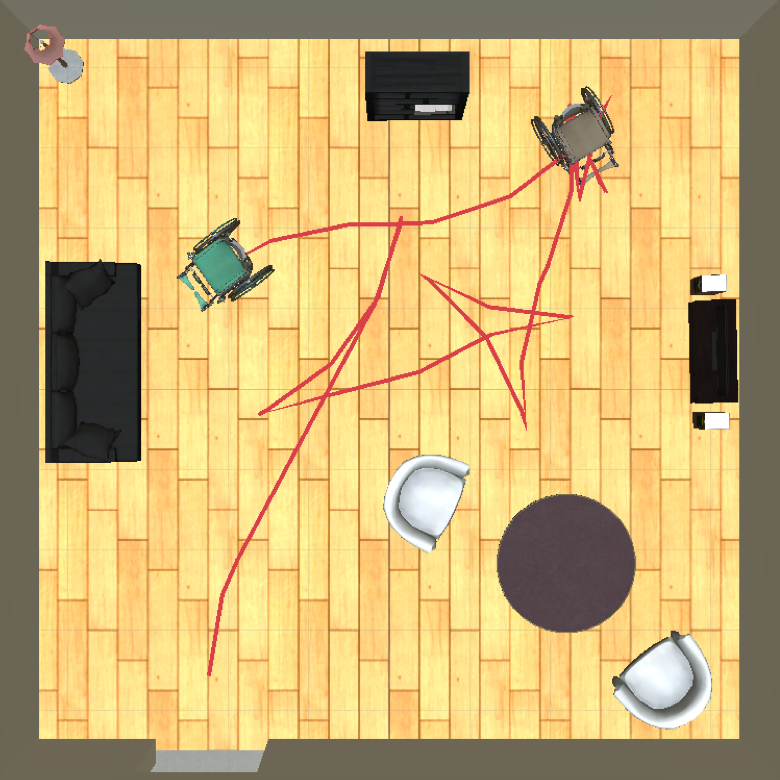}}

\caption{Four layout designs and a path covering three objects in each of the bedroom (top), office (mid),  and living room (bottom).  In all three examples, the wheelchair starts from door.}
\label{fig:cover3}
\label{fig:office}
\label{fig:living}
\end{figure*}

Motion planning takes on average 96.06\% of the total computation time of the optimization framework.
Naturally, we first show that the the proposed motion planning method based on the technique described in Sec.~\ref{sec:reuse-planner} is significantly faster  than  the baseline planner. 
Fig.~\ref{fig:time} shows the average path planning time  (in minutes) in three different environments.
The data is collected from 10 runs for each environment. 
As shown in the results, the proposed planner significantly outperforms the baseline planner in all cases. 
The proposed method reduces the planning time between 65\% to 70\%. 
Fig.~\ref{fig:cover3} shows four recommended layout designs for each room type. 
The trajectory shown for each design represents the tour of an objects sequence that has the largest probability  according to the given AO graph.

\subsection{Optimal Design in a Small Room}
\begin{figure}[ht]
{\includegraphics[width = 0.235\textwidth]{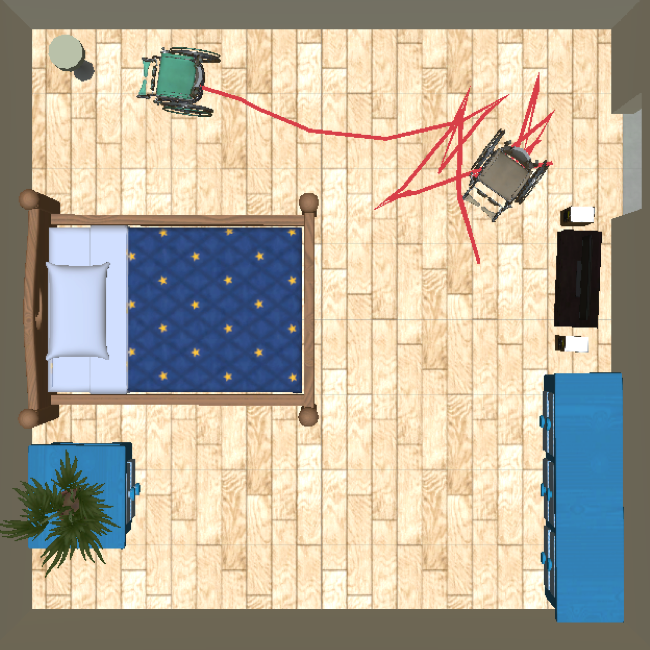}}
{\includegraphics[width = 0.235\textwidth]{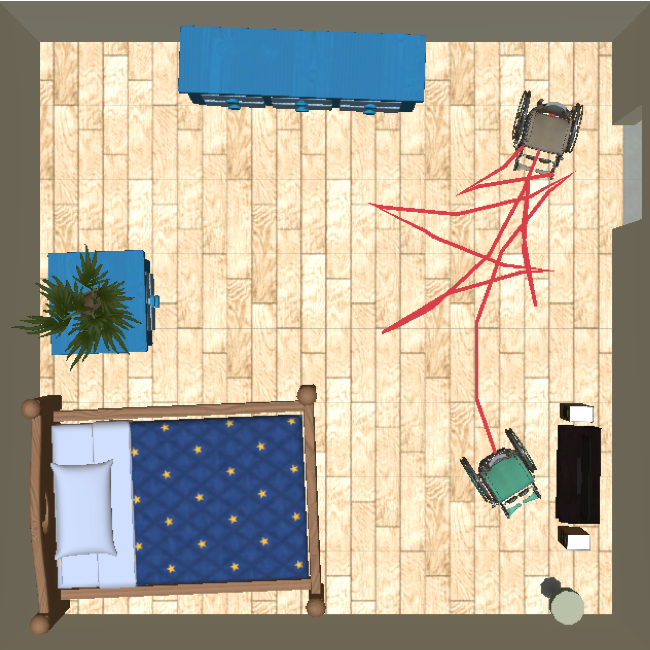}}
\caption{Two different layout designs for a small bedroom.  }
\label{fig:small}
\end{figure}

To test the proposed method in a smaller room, we reduce  the bedroom boundary length by 20\%, so the area of the new room is 36\% smaller. 
Fig.~\ref{fig:small} shows two designs of the smaller bedroom. The layouts are similar to those of the larger room. 
However, since the bedroom is more crowed, the motion planner takes much more time. 
The average path finding time in the small room is 30\% higher than that of the original bedroom. 

\subsection{Planner with different AO graphs}
\begin{figure}[ht]
{\includegraphics[width = 0.234\textwidth]{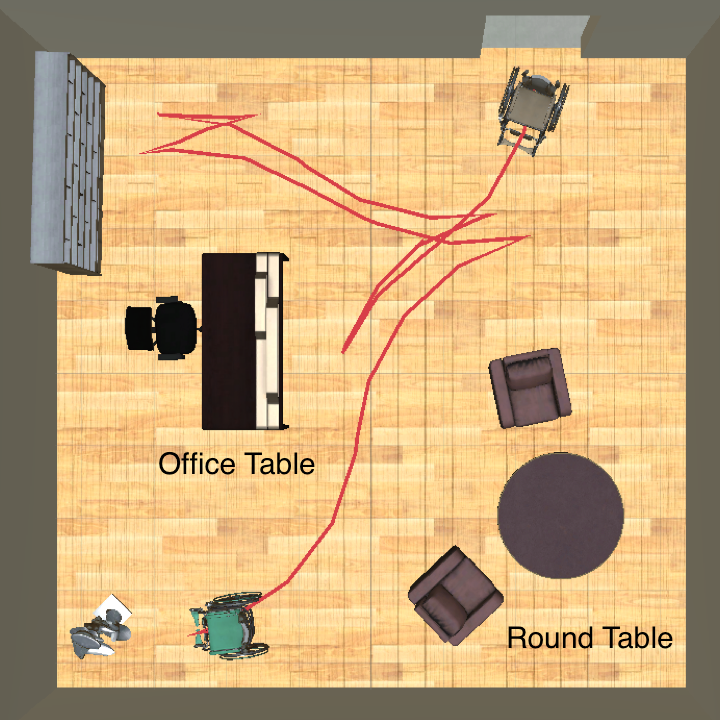}}
{\includegraphics[width = 0.234\textwidth]{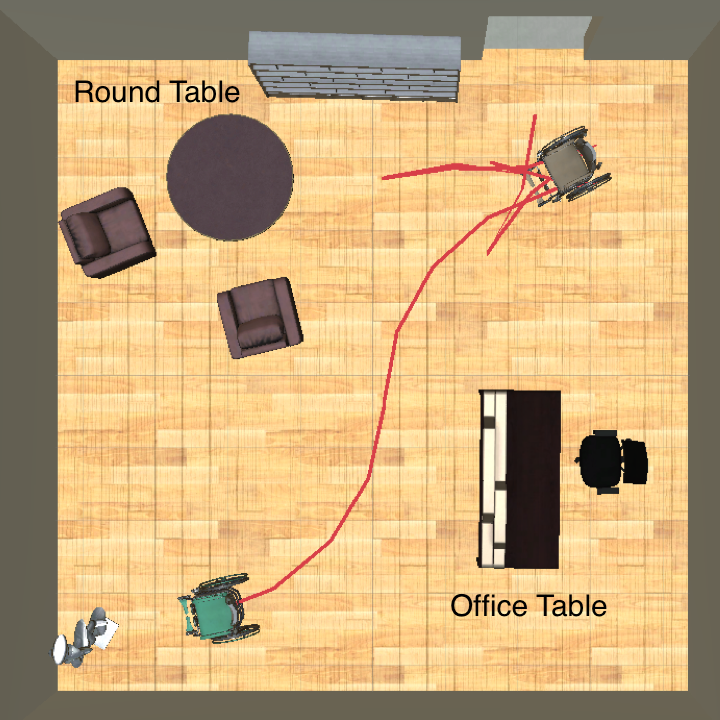}}
\caption{Two layout designs for the office with different AO graphs. Left: the office table has a larger weight than  the round table. Right: the round table has a larger path cost weight than the office table. }
\label{fig:actionoffice}
\end{figure}

Because our method considers accessibility in the scene synthesis,  
different user activities (encoded in the AO graphs)  should affect the designs of the human-robot shared space. 
Fig.~\ref{fig:actionoffice} shows two layouts created from  with two different AO graphs. 
For path cost weight, we set $\mathbf{w_P}=\alpha_i[1,1,1]$, where $\alpha_i$ is AO-graph dependent and  $[1,1,1]$ indicates that path length, rotation and clearance are equally important. 
These two AO graphs differ in the transition probabilities of the wheelchair moving (from other objects) to the office table and round table.
Because of these difference, recall the definition of the path cost weight in Section~\ref{sec:pathcost}, the scale factors $\alpha_i$ of the office table and round table consequently become different in these two examples. 
The AO graph used in the left figure  of Fig.~\ref{fig:actionoffice} results in a larger scaling factor thus a larger path cost weight of the office table than the weight of the round table, 
and the AO graph used in the right figure results in  a smaller scaling factor. We observe that the objects with larger weight are more likely to be placed in the 
area of the room that is more accessible from  the door. 




%
%
%
%
%
%

\section{Conclusion}
This paper presented the first known optimization framework for designing human-robot shared spaces by considering human preferences and  robot constraints. 
We also proposed a novel nonholonomic motion planner that can efficiently solve motion planning problems in many similar workspaces. 
The experiments showed that our framework can generate reasonable  layout designs, and the proposed motion planner takes significantly less time than the baseline planner. 
The experiments also showed that the method is consistent in creating similar designs for more crowded spaces, and
the differences in the AO graphs can be reflected in the synthesized layouts.

One limitation of our work is that the number and type of objects in the room are small. 
As the number of objects in the room increases, the number of possible sequences and tours increases exponentially. 
Further research is needed to overcome this.
The second limitation is that only rectangular rooms are considered. However, it should be straightforward to consider more complex room boundaries. 

\bibliographystyle{IEEEtran}
\bibliography{cite}


\end{document}